\documentclass{article}
\usepackage{spconf,amsmath,graphicx,hyperref}
\usepackage{amssymb}
\usepackage{multirow}

\title{Graph Neural Networks with Diversity-aware Neighbor Selection and Dynamic Multi-scale Fusion for Multivariate Time Series Forecasting}

\name{Jingqi Xu$^{1}$, Guibin Chen$^{2}$, Jingxi Lu$^{1}$, Yuzhang Lin$^{2}$\thanks{This material is based upon work supported in part by National Science Foundation (NSF) Award Number 2348289 and in part by the U.S. Department of Energy's Office of Energy Efficiency and Renewable Energy (EERE) under the Solar Energy Technologies Office Award Number 52834.}}

\address{
    $^{1}$University of Southern California \\
    $^{2}$New York University
}
\begin{document}
%
\maketitle
\begin{abstract}
Recently, numerous deep models have been proposed to enhance the performance of multivariate time series (MTS) forecasting. Among them, Graph Neural Networks (GNNs)-based methods have shown great potential due to their capability to explicitly model inter-variable dependencies. However, these methods often overlook the diversity of information among neighbors, which may lead to redundant information aggregation. In addition, their final prediction typically relies solely on the representation from a single temporal scale. To tackle these issues, we propose a Graph Neural Networks (GNNs) with Diversity-aware Neighbor Selection and Dynamic Multi-scale Fusion (DIMIGNN). DIMIGNN introduces a Diversity-aware Neighbor Selection Mechanism (DNSM) to ensure that each variable shares high informational similarity with its neighbors while maintaining diversity among neighbors themselves. Furthermore, a Dynamic Multi-Scale Fusion Module (DMFM) is introduced to dynamically adjust the contributions of prediction results from different temporal scales to the final forecasting result. 
Extensive experiments on real-world datasets demonstrate that DIMIGNN consistently outperforms prior methods.
\end{abstract}
\section{Introduction}

\label{sec:intro}
Multivariate Time Series (MTS) forecasting is a significant task with broad applications in various domains such as energy \cite{10758703,10761589}, 
finance \cite{PATTON2013899}, and transportation \cite{Jiang_2022}. 
\begin{figure}[t]
    \centering
    \includegraphics[width=0.6\linewidth]{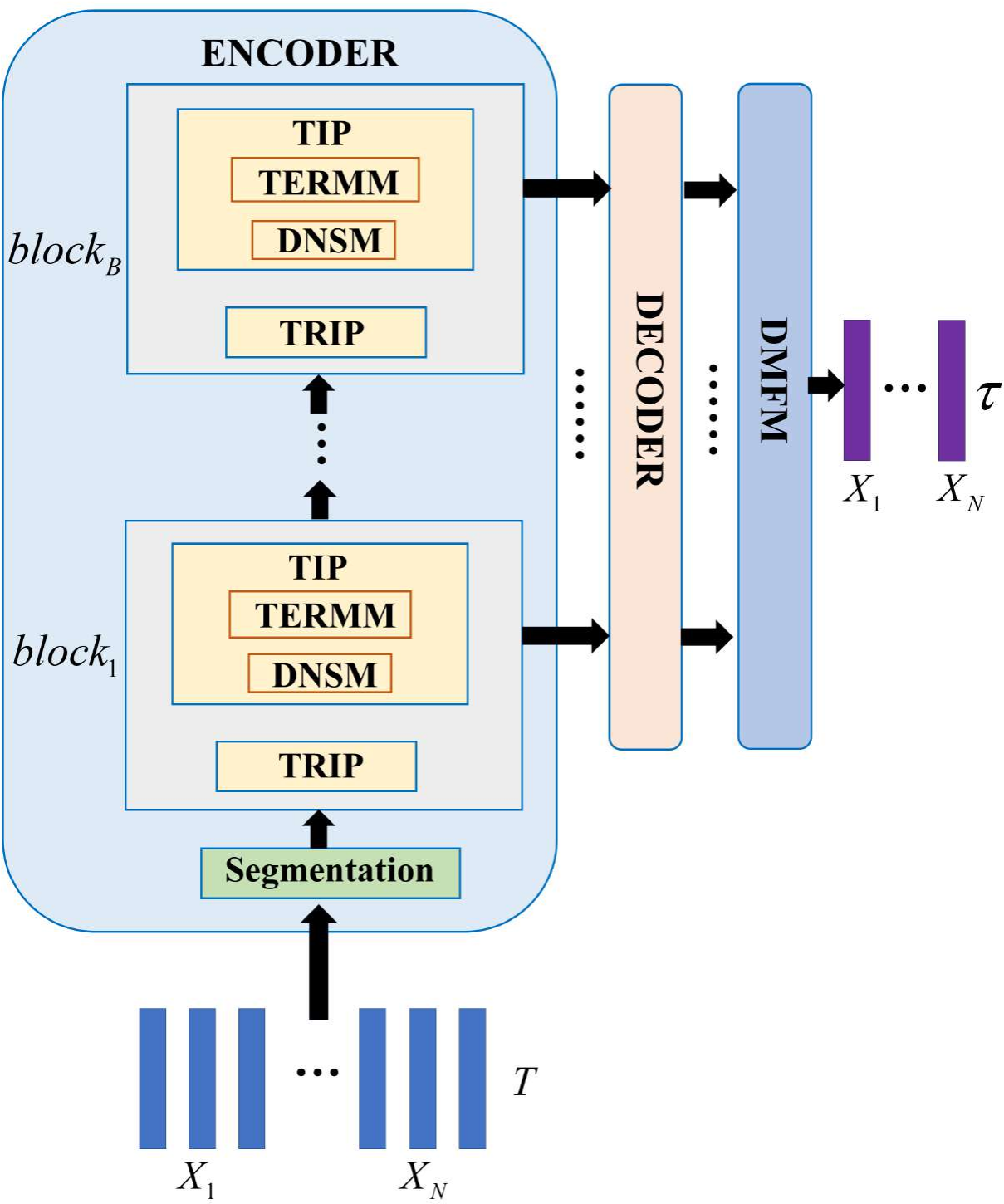}
    \caption{Overall architecture of DIMIGNN.}
    \label{fig:framework}
\end{figure}

Due to their strong ability to model nonlinear dependencies, deep learning methods \cite{11060930} have become the preferred choice for MTS forecasting. Recurrent neural networks (RNNs) \cite{schmidt2019recurrentneuralnetworksrnns} and their variants \cite{8053243} model temporal dependencies through recurrently connected hidden states. However, these models struggle to capture long-term temporal dependencies \cite{10448135}. 
Recently, several Transformer-based approaches \cite{vaswani2023attentionneed,zhou2021informerefficienttransformerlong,zhou2022fedformerfrequencyenhanceddecomposed,wu2022autoformerdecompositiontransformersautocorrelation} have been proposed, leveraging attention mechanisms \cite{vaswani2023attentionneed} to capture long-term dependencies. Nevertheless, they are unable to capture the complex inter-variable dependencies inherent in MTS data, which compromises their forecasting performance. Crossformer \cite{zhang2023crossformer} captures cross-time and cross-variable dependencies by leveraging a Two-Stage Attention (TSA) mechanism. However, it models spatiotemporal dependencies only for the target attribute of each variable, neglecting the importance of the remaining attributes in influencing the target attribute. For example, in renewable energy forecasting, the future power generation of a solar photovoltaic (PV) plant can be affected by various factors such as weather conditions and geographical location \cite{su142417005}.

Graph Neural Networks (GNNs) \cite{zhou2021graphneuralnetworksreview} based approaches represent a significant breakthrough in the field of MTS forecasting, as each variable can interact with its neighbor nodes to update its representation. 
HSDGNN \cite{ZHOU2025110304} captures temporal, inter-attribute, and inter-variable dependencies using two-level graph convolutions. However, existing GNNs-based methods do not explicitly consider the informational diversity of neighbors when selecting them for each variable, which may result in the aggregation of redundant information \cite{9457812}. Moreover, these methods overlook the fact that the predicted values are often influenced by representations at multiple temporal scales \cite{chen2023multiscaleadaptivegraphneural}.

To address the aforementioned challenges, in this paper, we propose a Graph Neural Networks with Diversity-aware Neighbor Selection and Dynamic Multi-scale Fusion (DIMIGNN) for MTS forecasting. DIMIGNN adopts an Encoder--Decoder--Fusion architecture. The encoder employs several stacked blocks to obtain representations at different temporal scales. In each block, we propose a Diversity-aware Neighbor Selection mechanism (DNSM) that selects neighbors for each variable through a scoring scheme that jointly considers information similarity and diversity, thereby explicitly ensuring that each variable shares high informational similarity with its neighbors while maintaining informational diversity among neighbors themselves. After the outputs from all encoder blocks are mapped by the decoder to prediction results at multiple temporal scales, a Dynamic Multi-Scale Fusion Module (DMFM) is introduced, which dynamically integrates these scale-specific predictions to produce the final prediction via a learnable weight vector.

Our contributions are summarized as follows. 
(1) We propose DIMIGNN, an innovative multi-scale temporal model for MTS forecasting.
(2) We introduce DNSM to ensure that each variable shares high informational similarity with its neighbors while maintaining informational diversity among neighbors themselves. 
(3) We present DMFM to dynamically adjust the contributions of forecasting results from different temporal scales to the final forecasting result.
(4) Extensive experiments on real-world datasets demonstrate that DIMIGNN achieves state-of-the-art performance.

\section{Proposed Method}
\label{sec:format}
\subsection{Problem Statement}

This section provides the mathematical formulation of the MTS forecasting task. 
Given the historical observations \(\mathbf{X}_{1:T} \in \mathbb{R}^{T \times N \times C}\), 
where \(T\) denotes the length of the past time series, \(N\) is the number of variables, 
and \(C\) is the number of attributes for each variable, we aim to predict the main attribute of each variable 
for the next \(\tau\) time steps in the future, denoted as \(\mathbf{X}_{T+1:T+\tau} \in \mathbb{R}^{\tau \times N\times1}\).
\subsection{Framework}
\begin{table*}[t]
\centering
\caption{Performance comparison of different models on \textbf{RPVPD} and \textbf{PSML}. Bold indicates the best performance. Results on \textbf{PSML} are 10 times larger than actual values for better readability}
\small
\setlength{\tabcolsep}{3.5pt}
\renewcommand{\arraystretch}{1.0}
\begin{tabular}{l@{\hspace{1pt}}c|cc|cc|cc|cc|cc|cc|cc}
\hline
\textbf{Models} & & \multicolumn{2}{c|}{\textbf{Informer}} & \multicolumn{2}{c|}{\textbf{Autoformer}} & \multicolumn{2}{c|}{\textbf{FEDformer}} & \multicolumn{2}{c|}{\textbf{LSTM}} & \multicolumn{2}{c|}{\textbf{HSDGNN}} & \multicolumn{2}{c|}{\textbf{ASTGCN}} & \multicolumn{2}{c}{\textbf{DIMIGNN}} \\
\textbf{Metric} & & MSE & MAE & MSE & MAE & MSE & MAE & MSE & MAE & MSE & MAE & MSE & MAE & MSE & MAE \\
\hline\hline
\multirow{5}{*}{\rotatebox{90}{\textbf{RPVPD}}}
& 1   & 0.180 & 0.216 & 0.126 & 0.216 & 0.129 & 0.223 & 0.183 & 0.220 & 0.162 & 0.211 & \textbf{0.112} & 0.198 & 0.113 & \textbf{0.169} \\
& 24  & 0.439 & 0.376 & 0.423 & 0.477 & 0.463 & 0.459 & 0.486 & 0.374 & 0.483 & 0.403 & 0.411 & 0.408 & \textbf{0.379} & \textbf{0.375} \\
& 48  & 0.504 & 0.416 & 0.483 & 0.525 & 0.565 & 0.509 & 0.597 & 0.438 & 0.461 & 0.417 & 0.529 & 0.458 & \textbf{0.434} & \textbf{0.404} \\
& 96  & 0.544 & 0.459 & 0.506 & 0.556 & 0.588 & 0.535 & 0.628 & 0.465 & 0.552 & 0.467 & 0.543 & 0.470 & \textbf{0.470} & \textbf{0.447} \\
& 168 & 0.592 & 0.473 & 0.547 & 0.567 & 0.623 & 0.543 & 0.567 & 0.450 & 0.514 & 0.464 & 0.594 & 0.484 & \textbf{0.480} & \textbf{0.444} \\
\hline
\multirow{5}{*}{\rotatebox{90}{\textbf{PSML}}}
& 1   & 0.370 & 1.510 & 0.980 & 2.450 & 0.008 & 0.204 & 1.552 & 3.839 & 0.005 & 0.044 & 0.005 & 0.056 & \textbf{0.003} & \textbf{0.036} \\
& 24  & 1.310 & 2.960 & 0.440 & 1.580 & 0.028 & 0.373 & 0.252 & 1.381 & 0.007 & 0.232 & 0.005 & 0.180 & \textbf{0.006} & \textbf{0.178} \\
& 48  & 1.220 & 2.820 & 0.680 & 1.950 & 0.052 & 0.502 & 0.110 & 0.874 & 0.007 & \textbf{0.162} & 0.009 & 0.185 & \textbf{0.006} & 0.169 \\
& 96  & 1.690 & 3.380 & 1.230 & 2.670 & 0.113 & 0.678 & 0.127 & 0.921 & 0.040 & 0.379 & 0.067 & 0.503 & \textbf{0.039} & \textbf{0.340} \\
& 168 & 1.630 & 3.190 & 2.380 & 3.701 & 0.134 & 0.695 & 0.189 & 1.006 & 0.157 & 0.773 & 0.405 & 1.256 & \textbf{0.113} & \textbf{0.663} \\
\hline
\textbf{Average} & & 0.848 & 1.580 & 0.780 & 1.469 & 0.270 & 0.472 & 0.469 & 0.997 & 0.239 & 0.355 & 0.268 & 0.420 & \textbf{0.204} & \textbf{0.323} \\
\hline
\end{tabular}
\label{tab:rpvpd-dsml-results}
\end{table*}


\begin{table}[t]
\centering
\footnotesize
\caption{Comparison of the performance of DIMIGNN with its variants}
\setlength{\tabcolsep}{2pt}
\renewcommand{\arraystretch}{1.1}
\begin{tabular}{l@{\hspace{4pt}}c|cc|cc|cc|cc}
\hline
\textbf{Models} & & \multicolumn{2}{c|}{\textbf{DIMIGNN}} & \multicolumn{2}{c|}{\textbf{(w/o)DyT}} & \multicolumn{2}{c|}{\textbf{(w/o)DNSM}} & \multicolumn{2}{c}{\textbf{(w/o)DMFM}} \\
\textbf{Metric} & & MSE & MAE & MSE & MAE & MSE & MAE & MSE & MAE \\
\hline\hline
\multirow{5}{*}{\rotatebox{90}{\textbf{RPVPD}}}
& 1   & \textbf{0.113} & \textbf{0.169} & 0.115 & 0.173 & 0.120 & 0.188 & 0.120 & 0.174 \\
& 24  & \textbf{0.379} & \textbf{0.375} & 0.386 & 0.388 & 0.390 & 0.457 & 0.402 & 0.420 \\
& 48  & 0.434 & \textbf{0.404} & \textbf{0.433} & 0.419 & 0.437 & 0.460 & 0.442 & 0.424 \\
& 96  & \textbf{0.470} & \textbf{0.447} & 0.475 & 0.460 & 0.475 & 0.509 & 0.474 & 0.456 \\
& 168 & \textbf{0.480} & \textbf{0.444} & 0.492 & 0.451 & 0.494 & 0.497 & 0.485 & 0.448 \\
\hline
\textbf{Average} & & \textbf{0.375} & \textbf{0.368} & 0.380 & 0.378 & 0.383 & 0.422 & 0.385 & 0.384 \\
\hline
\end{tabular}
\label{tab:ablation-results}
\end{table}

As shown in Figure~\ref{fig:framework},our method adopts an \linebreak[3] 
Encoder-Decoder-Fusion framework. The input matrix \(\mathbf{X}_{1:T}\) 
is first fed into the encoder, where it is processed by multiple blocks to progressively extract features with increasingly coarser temporal granularity. 
Specifically, the TRIP layer in each block merges adjacent segments produced by the lower block along the temporal dimension to obtain representations with coarser temporal scale, and then captures cross-time and cross-attribute dependencies via the multi-head self-attention mechanism. Based on the output of the TRIP layer, DNSM in the TIP layer scores each variable’s candidate neighbors using a scheme that jointly considers similarity and diversity, and selects the top-scoring candidates as the final neighbors. Using the neighbor matrix produced by DNSM, an attention-based Inter-Variable Merge Mechanism (TERMM) is applied to capture inter-variable dependencies and passes its output to the upper block and the decoder. The decoder utilizes the scale-specific output from each encoder block to generate predictions at different temporal scales, and DMFM then integrates these predictions using a learnable weight vector to obtain the final forecasting result.

\subsection{TRIP Layer}
\label{subsec:trip-layer}
To capture more informative patterns by aggregating nearby values along the temporal dimension, we follow the segmentation operation in  \cite{zhang2023crossformer} to segment and embed the input sequence \(\mathbf{X}_{1:T}\) along the time axis. This yields a set of embeddings 
\(H = \{ h_{i,d,c} \mid 1 \leq i \leq \frac{T}{L_s}, 1 \leq d \leq D, 1 \leq c \leq C \}\), 
where \(L_s\) is the segment length, and \(h_{i,d,c} \in \mathbb{R}^{d^{\text{hidden}}}\) denotes the embedding of the \(c\)-th attribute of the \(d\)-th variable in the \(i\)-th time segment.

We denote the input to the TRIP Layer as \(Z \in \mathbb{R}^{L \times D \times C \times d_{\text{hidden}}}\), where \(Z\) can be either \(H\) or the output \(Y\) from the lower encoder block. Inspired by results in computer vision and natural language processing showing that replacing LayerNorm with Dynamic Tanh (DyT) yields better performance ~\cite{zhu2025transformersnormalization}, we apply a DyT-based multi-head self-attention mechanism to capture cross-time and cross-attribute dependencies as follows:
\begin{equation}
\left\{
\begin{aligned}
Z = H\!:&\quad \hat{Z}^{\text{time}} = \text{DyT}(Z + \text{MSA}^{\text{time}}(Z, Z, Z)) \\
Z = Y\!: &\quad \hat{Z}^{\text{time}} = \text{DyT}(Z + \text{MSA}^{\text{time}}(\text{M}(Z), \text{M}(Z), \text{M}(Z)))
\end{aligned}
\right.
\end{equation}

\vspace{-1.1em}
\begin{equation}
Z^{\text{time}} = \text{DyT}(\hat{Z}^{\text{time}} + \text{MLP}(\hat{Z}^{\text{time}}))
\end{equation}

\vspace{-1.1em}
\begin{equation}
\hat{Z}^{\text{attr}} = \text{DyT}(Z^{\text{time}} + \text{MSA}^{\text{attr}}(Z^{\text{time}}, Z^{\text{time}}, Z^{\text{time}}))
\end{equation}

\vspace{-1.1em}
\begin{equation}
Z^{\text{attr}} = \text{DyT}(\hat{Z}^{\text{attr}} + \text{MLP}(\hat{Z}^{\text{attr}}))
\end{equation}
where \(\mathrm{DyT}(\cdot)\) is defined as \(\mathrm{DyT}(x)=\gamma\,\tanh(\alpha x)+\beta\), with \(\alpha\) a learnable scalar and \(\gamma,\beta\) learnable vectors. The operator \(M(\cdot)\) denotes temporal segment merging, which aggregates adjacent segments along the temporal dimension to obtain coarser-grained representations ~\cite{zhang2023crossformer}. \(\mathrm{MSA}^{\text{time}}(\cdot)\) and \(\mathrm{MSA}^{\text{attr}}(\cdot)\) denote multi-head self-attention applied along the temporal and attribute dimensions, respectively.
\subsection{TIP Layer}
In the TIP Layer, we first employ our proposed DNSM to select $K$ neighbors for each variable. Subsequently, TERMM is applied to effectively capture cross-variable dependencies.

1) DNSM:The output of the TRIP layer \( Z^{\text{attr}} \) is averaged along the temporal dimension to obtain \( Y^{\text{attr}} = \{ Y^{\text{attr}}_i \mid 1 \leq i \leq N \} \), where each \( Y^{\text{attr}}_i \in \mathbb{R}^C \) denotes the temporal average representation of the \( i \)-th variable.

To ensure that each variable shares high informational similarity with its neighbors while maintaining informational diversity among the neighbors themselves, we propose a scoring scheme. Specifically, for each variable \(i \in \{1,\ldots,N\}\), we first select, from the remaining variables, the one with the highest cosine similarity to the variable \(i\) as the first neighbor, denoted \(j_1\):
\begin{equation}
j_1 = \arg\max_{j \ne i} \frac{Y^{\text{attr}}_i \cdot Y^{\text{attr}}_j}{\| Y^{\text{attr}}_i \| \cdot \| Y^{\text{attr}}_j \|}
\end{equation}
Then from the remaining unselected variables, we iteratively select the \( m \)-th neighbor \( j_m \):
\begin{equation}
s_{\mathrm{sim}}^{\,j}
= \frac{Y^{\text{attr}}_{i}\!\cdot Y^{\text{attr}}_{j}}
       {\|Y^{\text{attr}}_{i}\|\cdot \|Y^{\text{attr}}_{j}\|}
\end{equation}

\begin{equation}
s_{\mathrm{div}}^{\,j}
= 1-\frac{1}{m-1}\sum_{n=1}^{m-1}
  \frac{Y^{\text{attr}}_{j}\!\cdot Y^{\text{attr}}_{j_n}}
       {\|Y^{\text{attr}}_{j}\|\cdot \|Y^{\text{attr}}_{j_n}\|}
\end{equation}

\begin{equation}
j_m=\arg\max_{j\in S}\big[\lambda\,s_{\mathrm{sim}}^{\,j}+(1-\lambda)\,s_{\mathrm{div}}^{\,j}\big]
\end{equation}
where \(S=\{\, j \neq i \mid j \notin \{j_1,\ldots,j_{m-1}\}\,\}\) denotes the set which consists of the current candidate variables. For each candidate \(j\in S\), \(s_{\mathrm{sim}}^{\,j}\) is the similarity score that measures its similarity to the variable \(i\), whereas \(s_{\mathrm{div}}^{\,j}\) is the diversity score that quantifies its dissimilarity to neighbors of the variable \(i\) already selected.  \(\lambda\in[0,1]\) controls the balance between \(s_{\mathrm{sim}}^{\,j}\) and \(s_{\mathrm{div}}^{\,j}\), thereby explicitly encouraging the variable \(i\) to share high informational similarity with its neighbors while maintaining informational diversity among the neighbors themselves. After selecting \(k\) neighbors for each variable \(i \in \{1,\ldots,N\}\), we form the neighbor matrix \(N=\{N_i \mid 1\le i\le N\}\), where \(N_i\in\mathbb{R}^{k}\) represents the \(k\) neighbors of the \(i\)-th variable. In our experiments (Section~\ref{sec:pagestyle}), we empirically set \(\lambda=0.7\) and \(k=3\).

2) TERMM: We adopt a method similar to the previous work \cite{veličković2018graphattentionnetworks} to capture inter-variable dependencies given the neighbor matrix \( N \):
\begin{equation}
Z^{\text{va}} = GAT(Z^{\text{attr}},N)
\end{equation}
where \( GAT(\cdot) \) denotes the graph attention network.

Therefore, the complete process of each encoder block is:
\begin{equation}
Y = Z^{\text{va}} = TIP(TRIP(Z)).
\end{equation}
where \(Y\) is the output of each block, which is passed to the upper block and the decoder.
\subsection{DMFM}
We adopt a decoder similar to \cite{zhang2023crossformer}, which utilizes the outputs from each block in the encoder to generate multiple prediction results, denoted as \( \hat{X}_{T+1:T+\tau}^{\text{pred}} = \left\{ \hat{X}_{T+1:T+\tau}^{\text{pred}, b} \mid 1 \leq b \leq B \right\} \), where \( B \) is the number of blocks in the encoder, and \( \hat{X}_{T+1:T+\tau}^{\text{pred}, b} \in \mathbb{R}^{\tau \times N\times1} \) represents the prediction result from the \( b \)-th block. Since the merge operation \( M(\cdot) \) in the TRIP Layer merges two adjacent temporal segments into a coarser-grained representation along the temporal dimension, the final prediction \( \hat{X}_{T+1:T+\tau}^{\text{pred}} \) contains forecasting values at multiple temporal scales.We propose DMFM to dynamically fuse forecasting values at multiple temporal granularities to obtain the final prediction \(X_{T+1:T+\tau}^{\text{pred}}\):
\begin{equation}
\begin{aligned}
X_{\text{ave}}^{\text{pred}} &= \frac{1}{B} \sum_{b=1}^{B} \hat{X}_{T+1:T+\tau}^{\text{pred}, b} \\
\bar{\alpha} &= \text{ReLU}(W_1 X_{\text{ave}}^{\text{pred}} + b_1) \\
\alpha &= \text{softmax}(W_2 \bar{\alpha} + b_2) \\
{X}_{T+1:T+\tau}^{\text{pred}} &= \sum_{b=1}^{B} \alpha[b] \hat{X}_{T+1:T+\tau}^{\text{pred}, b}
\end{aligned}
\end{equation}
where \( W_1 \) and \( W_2 \) are learnable weight matrices, \( b_1 \) and \( b_2 \) are learnable bias vectors. \( \alpha \in \mathbb{R}^B \) denotes the weight vector for forecasting values at multiple temporal scales.
\section{EXPERIMENTS}
\label{sec:pagestyle}

In this section, we adopt Mean Squared Error (MSE) and Mean Absolute Error (MAE) as evaluation metrics to assess the performance of our proposed DIMIGNN model. For comparison, we consider the following models: Informer \cite{zhou2021informerefficienttransformerlong}, Autoformer \cite{wu2022autoformerdecompositiontransformersautocorrelation}, FEDformer \cite{zhou2022fedformerfrequencyenhanceddecomposed}, LSTM \cite{6795963}, HSDGNN \cite{ZHOU2025110304}, and ASTGCN \cite{liu2023attentionbasedspatialtemporalgraphconvolutional}.
\begin{figure}[t]
    \centering
    \includegraphics[width=0.78\linewidth]{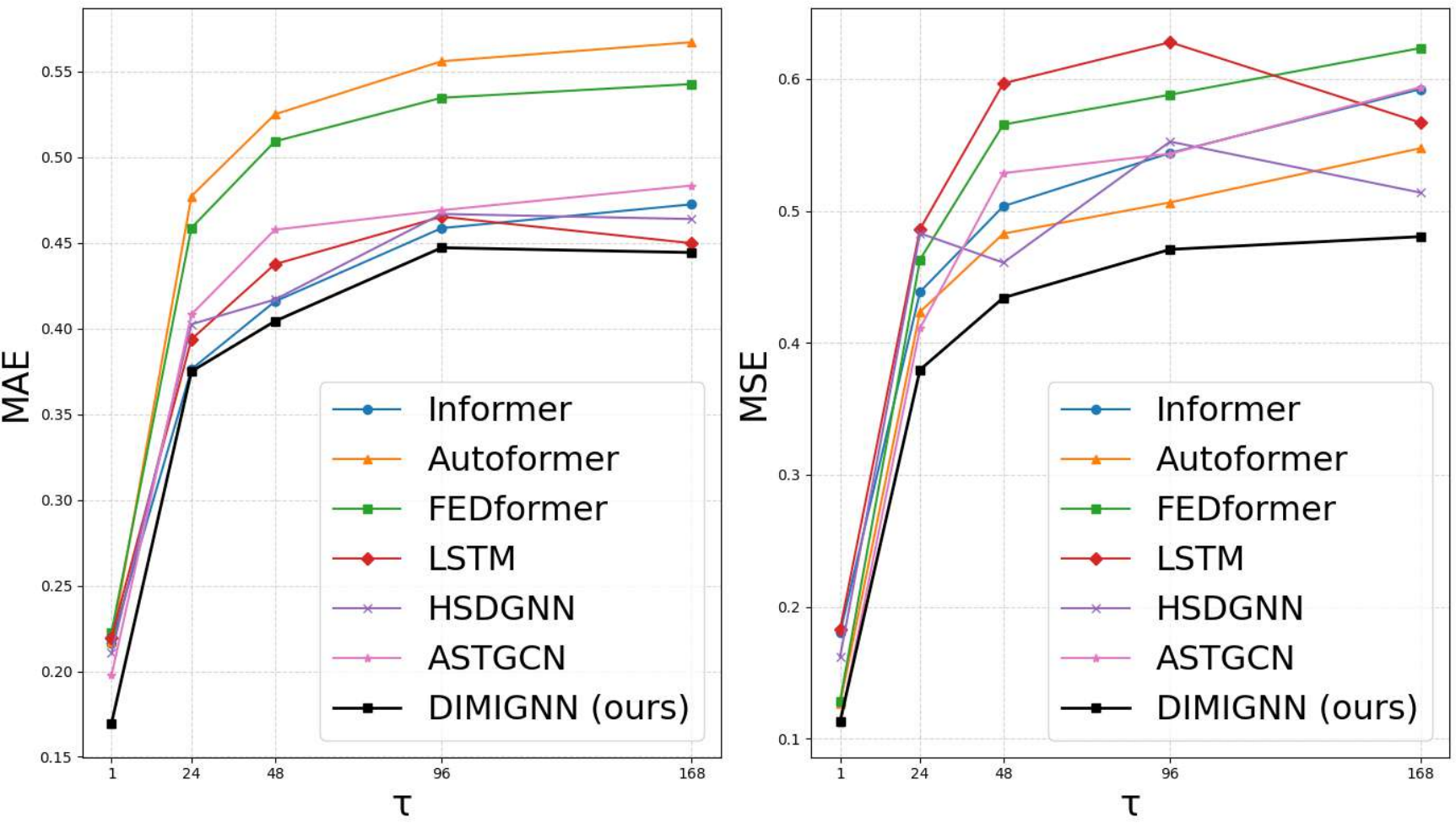}
    \caption{The variation of MAE and MSE with different prediction lengths on \textbf{RPVPD}.}
    \label{fig:rpvpd-error}
\end{figure}
\subsection{Datasets}
We conduct experiments on two publicly available real-world datasets: Regional PV Power Dataset \textbf{RPVPD} \cite{10758703} and Electricity dataset \textbf{PSML} \cite{ZHOU2025110304}. \textbf{RPVPD} contains the hourly solar PV power and 7 meteorological features of 127 distributed PV plants in the New England region of the United States from 00:00, July 1, 2019 to 23:00, March 31, 2022. 
\textbf{PSML} contains the minute-level load power and 3 features of 66 regions in the United States from 00:00, January 1, 2018 to 23:59, January 31, 2018. 
We set the PV power and load power as the target attributes for \textbf{RPVPD} and \textbf{PSML}, respectively. we allocate 70\% for training, 10\% for validation, and 20\% for testing.

\subsection{Experimental Details}
All models follow the same experimental settings. To evaluate the short-term and long-term forecasting capabilities of our method, we set the prediction length $\tau$ as $\tau \in \{1, 24, 48, 96, 168\}$. We use ADAM as the optimizer, and all experiments are performed on NVIDIA A40 GPUs.
\subsection{Experimental Results}
Table~\ref{tab:rpvpd-dsml-results} show the forecasting results of our method and established methods on \textbf{RPVPD} and \textbf{PSML} across different prediction lengths. Figure~\ref{fig:rpvpd-error} illustrates the changes in forecasting errors with increasing prediction lengths on \textbf{RPVPD}. Table~\ref{tab:rpvpd-dsml-results} clearly indicates that our method consistently outperforms established methods across all prediction lengths, underlining its effectiveness in both short-term and long-term forecasting. Specifically, compared with the best-performing established method HSDGNN, our method achieves average improvements of 14.6\% in MSE and 9.0\% in MAE. As can be seen from Figure~\ref{fig:rpvpd-error}, graph-based methods generally outperform non-graph-based methods at different prediction lengths, highlighting the importance of capturing inter-variable dependencies in MTS forecasting. Moreover, the growth rate of forecasting error with increasing prediction length is lower for our method compared to most of other methods, further demonstrating its effectiveness, especially in long-term forecasting scenarios.

\subsection{Ablation Study}
We conduct an ablation study on the key components of our method using \textbf{RPVPD}. 
Specifically, DIMIGNN without DyT refers to replacing DyT with LayerNorm, 
DIMIGNN without DNSM indicates that only similarity is considered when selecting neighbors 
for each variable, i.e., $\lambda = 1$ in Equation (6), and 
DIMIGNN without DMFM denotes directly summing the predictions at different temporal scales. 
The test results are presented in Table~\ref{tab:ablation-results}. 
It can be observed that removing any component leads to a decrease in both short-term and long-term 
forecasting accuracy. These results highlight the importance of each component in DIMIGNN.
\section{Conclusion}
In this work, we introduce DIMIGNN for MTS forecasting. 
Specifically, 
we propose DNSM to ensure that the selected neighbors for each variable exhibit 
informational diversity. In addition, DMFM is introduced, which dynamically integrates predictions at different time scales using a learnable weight vector to generate the final forecasting result. 
Experimental results on real-world datasets demonstrate the effectiveness of DIMIGNN
compared to established methods.
\bibliographystyle{IEEEbib}
\bibliography{strings,refs}

\end{document}